# Towards AI-Assisted Generation of Military Training Scenarios

**Soham Hans, Volkan Ustun, Mark Core, Benjamin Nye**

USC Institute for Creative Technologies

Playa Vista, CA

sohamhan@usc.edu, ustun@ict.usc.edu, core@ict.usc.edu, nye@ict.usc.edu

**James Sterrett, Matthew Green**

U.S. Army University
Command and General Staff College

Fort Leavenworth, KS

james.j.sterrett.civ@army.mil, matthew.k.green.civ@army.mil

## ABSTRACT

Achieving expert-level performance in simulation-based training relies on the creation of complex, adaptable scenarios—a traditionally laborious and resource-intensive process. Although prior research explored scenario generation for military training, pre-LLM AI tools struggled to generate sufficiently complex or adaptable scenarios. This paper introduces a multi-agent, multi-modal reasoning framework that leverages Large Language Models (LLMs) to generate critical training artifacts, such as Operations Orders (OPORDs). Our approach uses AI to automate and enhance scenario generation tasks, significantly reducing manual workload and accelerating development. We structure our framework by decomposing scenario generation into a hierarchy of subproblems, and for each one, defining the role of the AI tool: (1) generating options for a human author to select from, (2) producing a candidate product for human approval or modification, or (3) generating textual artifacts fully automatically. Our framework employs specialized LLM-based agents to address distinct subproblems. Each agent receives input from preceding subproblem agents — integrating both text-based scenario details and visual information (e.g., map features, unit positions) — and applies specialized reasoning to produce appropriate outputs. Subsequent agents process these outputs sequentially, preserving logical consistency and ensuring accurate document generation. This multi-agent strategy overcomes the limitations of basic prompting or single-agent approaches when tackling such highly complex tasks. We validate our framework through a proof-of-concept that generates the scheme of maneuver and movement section of an OPORD while estimating map positions and movements as a precursor — demonstrating its feasibility and accuracy. Our results demonstrate the potential of LLM-driven multi-agent systems to generate coherent, nuanced documents and adapt dynamically to changing conditions — advancing automation in scenario generation for military training.

## ABOUT THE AUTHORS

**Soham Hans** is a Research Project Specialist with the Human-Inspired Adaptive Teaming Systems Group at the University of Southern California's Institute for Creative Technologies. His work focuses on applying multimodal generative AI technologies to create scenario-based environments that support reinforcement learning. He also explores the use of generative language models for cognitive analysis in the context of cybersecurity.

**Volkan Ustun, Ph.D.** is the Associate Director of the Human-Inspired Adaptive Teaming Systems Group at the USC Institute for Creative Technologies. His research augments Multi-agent Reinforcement Learning (MARL) with generative models, drawing inspiration from operations research, human judgment and decision-making, game theory, graph theory, and cognitive architectures to better address the challenges of developing behavior models for synthetic characters, mainly in military training simulations.

**Benjamin Nye, Ph.D.** is Director of Learning Science at the USC Institute for Creative Technologies (ICT). Dr. Nye's research has been recognized for excellence in intelligent tutoring systems (First Place ONR ITS STEM Grand Challenge), cognitive agents (BRIMS 2012 best paper), realistic behavior in training simulations (Federal Virtual Worlds Challenge), and machine learning for adaptive systems (OMEGA, I/ITSEC 2021 Best Paper Overall, with Eduworks). Nye's research is on scalable learning technologies and design principles that promote learning. His recent work emphasizes AI tools for instructors and content developers to use AI tools that enable them to rapidly update content and to create AI-enabled learning experiences (i.e., AI-human teams to generate AI tutoring).

**Mark G. Core, Ph.D.** is a Research Scientist in ICT's Learning Science, researching topics such as authoring tools, natural language processing, virtual reality and data analytics. He has over 15 years of experience in developing and evaluating virtual role players for learning, such as BiLAT (training of bilateral negotiation, winner of a 2008 Army Modeling and Simulation Award), and the Standard Patient Studio (training of medical interviewing, winner of multiple awards including 2016 Best Government Game, I/ITSEC). He has also published research on machine learning pipelines for automated analyses, including frameworks such as SLATS (Semi-Supervised Learning for Assessment of Teams in Simulations) and RACR (Rapid Adaptive Content Registry).

**LTC(R) Matthew Green** is the Scenario Development Team Leader at the United States Army Command and General Staff School (CGSS), Army University. He has 20 years of active-duty military experience serving in Armor and Cavalry units and on Corps planning staffs. As a Department of the Army civilian, he has an additional 15 years of experience teaching tactics, planning, and decision making in the Department of Army Tactics at CGSS. His team has developed multiple training scenarios to support the CGSS curriculum to include the PACIG AEGIS family of orders used in this body of work.

**James Sterrett, Ph.D.** is the Chief of the Simulation Education Division in the Directorate of Simulation Education of U.S. Army University/Command & General Staff College. Since 2004, he has taught the use and design of simulations and games and supported their use in education. He also earned a PhD in War Studies from King's College London and has participated in beta test and design teams for many games, notably including Steel Beasts and Attack Vector: Tactical.

# Towards AI-Assisted Generation of Military Training Scenarios

**Soham Hans, Volkan Ustun, Mark Core, Benjamin Nye**

USC Institute for Creative Technologies

Playa Vista, CA

sohamhan@usc.edu, ustun@ict.usc.edu, core@ict.usc.edu, nye@ict.usc.edu

**James Sterrett, Matthew Green**

U.S. Army University
Command and General Staff College

Fort Leavenworth, KS

james.j.sterrett.civ@army.mil, matthew.k.green.civ@army.mil

## INTRODUCTION

Achieving expertise through deliberate practice is hindered by the complexity and cost of building realistic training scenarios. Creating high-fidelity live, virtual, and constructive scenarios is expensive and requires specialized expertise, limiting both their number and diversity. This is especially relevant to efforts such as the Army's Synthetic Training Environment (STE), which aims to broaden Soldiers' experiential learning: the need to adapt those scenarios to new terrain, weather, and opposing-force objectives further compound the problem.

Earlier research has explored automated scenario generation, but the Artificial Intelligence (AI) and Machine Learning (ML) tools historically could not produce scenarios that were both complex and adaptable. Some systems varied only a few parameters optimized with reinforcement learning (Rowe et al., 2018), while others combined cognitive-task-analysis models with novelty search (Dargue et al., 2019; Folsom-Kovarik et al., 2019). These methods struggled to transfer to new contexts and to achieve the realism demanded by advanced training. Current large language models (LLMs) promise major progress. By drawing on deep data sets —including historical operations, current geopolitical contexts, and doctrinal publications— LLMs can create context-rich scenarios that adapt dynamically to changing constraints. The military is actively exploring this potential (Caballero et al., 2024). Proposed benefits include increasing the realism of virtual training environments by generating simulated social media feeds that reflect in-scenario events and creating enhanced 'pattern of life' entities (Hill, 2024).

However, swapping a full scenario-authoring pipeline for a single LLM would not realistically reflect the collaborative, role-specialized processes and review that produce training scenarios designed to practice specific competencies. We therefore propose a multi-agent reasoning approach that distributes cognitive effort across specialized LLM agents that will effectively support the scenario developers, establishing a human-AI scenario co-generation framework. This approach mirrors effective human teamwork (i.e., agents focus on specific tasks with organizational structures guiding the sharing of information) to maintain coherence in high-dimensional problem spaces and overcome the scaling limitations of single-agent prompting.

To build our LLM-based multi-agent reasoning framework for scenario generation, the team analyzed the process employed by Army University experts in building scenarios for large scale training exercises. This analysis led to a decomposition of the scenario generation process into a hierarchy of subproblems that is further discussed in the Subproblem Hierarchy section. Furthermore, for each subproblem in this hierarchy, we defined the role that our proposed framework will take: (1) generating options for a human author to select from, (2) producing a candidate product for human approval or modification, or (3) generating textual artifacts fully automatically. This approach was applied to the Pacific Pugilist training from the Army University Pacific Aegis scenario, and demonstrated our framework's efficacy in sample subproblems for Pacific Pugilist.

## SCENARIO GENERATION FOR LARGE SCALE TRAINING EXERCISES

Army University maintains a small scenario development team that creates fictional conflicts set in TRADOC G2's Decisive Action Training Environment (DATE) framework. DATE includes models for multiple regions, including Eurasia, and the Indo-Pacific. From those general environments, the scenario writers create specific problem sets tailored to the learning objectives of a supported professional military education (PME) course. In a very simple overview of the process flow, the scenarios start with the analysis of those learning objectives. They determine the appropriate military echelon (Corps, Division, Brigade) the learner will be representing, the kinds of tasks or operations they need to be able to do (for example: river crossing, passage of lines, deliberate defense), and the kinds of organizations they need to do it with (armored brigade, airborne division, marine regiment). From there, they expand on that organization and build out the friendly task organization two echelons up, two echelons down, and laterally to include appropriate adjacent and supporting units. With the total scope of the friendly forces defined, the focus shifts to developing the threat order of battle and tasks that are required to stimulate the learning objectives. Next, the focus shifts to finding appropriate terrain for those two forces to engage each other on. For example, finding a location with an appropriate river, if a river crossing is required. After finding an appropriate terrain, scenario authors begin work on the road to war. This is a plausible sequence of events that lead up to the point in time where the student engages with the problem. The road to war gives context, logic, and purpose to the problem. This can take several iterations to get correct, especially if the curriculum requires several problems to be solved in a linked series of exercises. For example, a deployment into theater, followed by a defense, followed by a transition to the offense. Once all of this has been established, scenario authors then express the output in doctrinal documents such as Warning Orders, Operations Orders, Fragmentary Orders, and intelligence summaries. They also produce any necessary graphical overlays, databases, and simulation support products. Traditionally, these products reflect realistic scenarios but are not set in specific real-life political contexts (e.g., fictional countries, but with geography or conflict types relevant to different regions of the world). These products are then maintained over time to reflect changes in doctrine, organizations, equipment, and learning objectives.

In this work, we use the PACIFIC AEGIS scenario as a gold-standard, human-authored reference point. PACIFIC AEGIS is a name of a scenario available for use in PME across TRADOC. It is set in the DATE-Indopacific operational environment. The scenario consists of a nested family of orders starting with the Joint Task Force (JTF) and flowing down through the Land Component Command (LCC), tactical corps, division, and finally the brigade echelon. The targeted audience for the product is the Army division. To simulate students acting as a division staff they are given Corps level Operations Orders (OPORD) as the basis of their analysis and planning. The products include three OPORDs. PACIFIC PASSAGE simulates unit deployment, PACIFIC PARAPET focuses on defensive operations, and PACIFIC PUGILIST is designed to support offense operations. In each case the trainee receives the order from the I Corps and acts as staff of the 25th Infantry division. PACIFIC PUGILIST also includes two other ORDERS. APOLLO is a division level product based on the 25th ID's mission. ADONIS is a brigade level OPORD derived from one of the brigades of the 25th. Collectively, the trainee can determine how each echelon of the force receives instructions from its higher formation.

**DEPENDENCY HIERARCHY FOR SCENARIO GENERATION**

Building on the expert-driven scenario development process described above, we formalize a modular structure that decomposes the scenario into discrete, interdependent information blocks. Each block represents a specific set of information relevant to the scenario. By analyzing dependencies between blocks, we can define a directed graph that guides the order in which these blocks should be constructed, as shown in Fig. 1. This structure supports a clearer understanding of input-output relationships between blocks and allows for the targeted automation of individual components within the broader scenario generation process.

By isolating key elements in this modular way, we can systematically assess what information is required to generate each block, paving the way for incremental automation. Below, we describe some of the primary information blocks and their roles:

- **Backstory**: Captures pre-conflict conditions and political/military developments leading up to the scenario, typically derived from the Road to War.
- **Learning Objectives**: Defines what military competencies or knowledge areas the training exercise is designed to develop.
- **Map and MCOO (Modified Combined Obstacle Overlay)**: Includes terrain data, phase lines, avenues of approach, mobility corridors, and key terrain features that influence maneuver planning.
- **Force Groupings**: Specifies the friendly and opposing units, organized across echelons (e.g., divisions, brigades, battalions).

- **Red and Blue Objectives**: Describes the high-level goals and end-states of the opposing forces in the simulated conflict.
- **High-Level Unit Purpose**: Articulates each unit's assigned tasks and the rationale behind those tasks.
- **Decision Support Matrix**: Encodes planned responses to triggers or contingencies, enabling dynamic adaptation during execution.
- **Time-Based Unit Positions**: Details the geographic locations and boundaries of units over time, including reference to phase lines and control measures.

Once these foundational blocks are generated, they serve as inputs to produce formal doctrinal documents such as Operations Orders (OPORDs). Since OPORDs adhere to a standardized structure with multiple sections and subsections, we can map each of these paragraphs back to specific information blocks. This idea is used to broaden the dependency information flow graph. This alignment ensures coherence across the document by grounding each part in a shared set of consistent source materials.

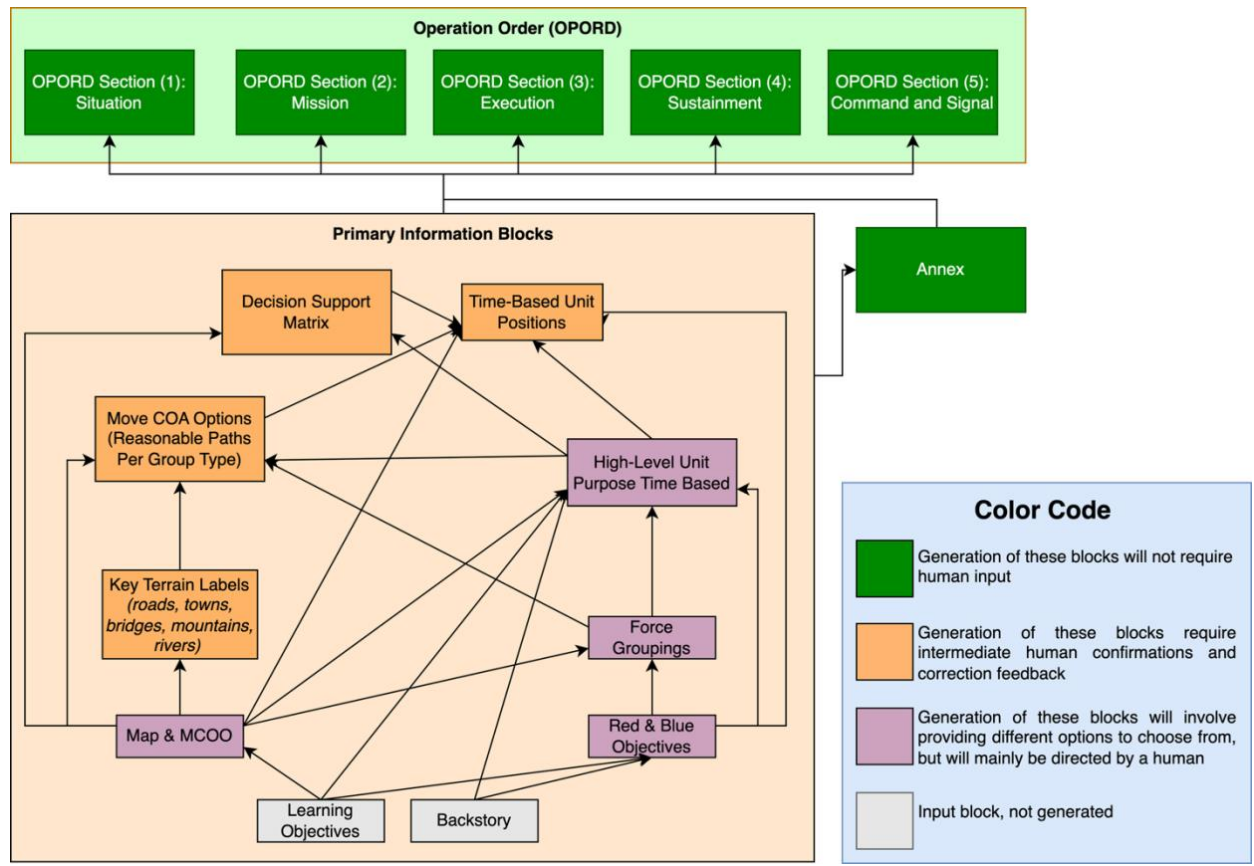

**Fig 1. Information Flow Graph representing building blocks of how the learning objectives and backstory can be expanded step by step into concepts that help develop the rest of the scenario**

To guide automation, we categorize each information block based on the level of human involvement required for its generation. **Green blocks** are fully automatable, assuming accurate inputs from upstream blocks. These tasks typically involve reformatting information or generating standardized language, both of which fall well within the capabilities of current LLMs. **Orange blocks** can also be handled by LLMs, but they require human verification. These blocks often involve more complex reasoning, where expert judgment may be necessary to validate the output or provide corrective guidance. **Purple blocks**, by contrast, are best managed by human experts due to the need for subjective decision-making or creative judgment. In such cases, the LLM plays a supportive role—generating options, suggesting directions, or assisting in iterative ideation with the human in the loop.

In our proof-of-concept example, we focus on the generation of two information blocks and demonstrate the system's ability to both support expert-in-the-loop workflows and autonomous generation of complex doctrinal content.

- **Time-Based Unit Positions**, classified as an Orange Block, showcases how an LLM-based framework can perform complex spatial and temporal reasoning. This block is especially useful for demonstrating human-AI collaboration, as the outputs are easily verifiable by experts, who can approve or iteratively guide the system when its reasoning diverges from expectations.
- **OPORD Subsection: Scheme of Movement and Maneuver**, a Green Block, was selected because it is one of the most difficult sections categorized as automatable. This section requires transforming spatial reasoning about force movements into doctrinal language—a task that has historically challenged LLMs. By targeting this block, we aim to highlight that even high-difficulty components within the automatable space can be addressed using structured frameworks and scenario-specific prompts.

**MULTI-AGENT, MULTI-MODAL GENERATIVE AI FRAMEWORK**

To support the automated generation of information blocks described earlier, we developed a multi-agent framework based composed of large language model (LLM) agents with distinct roles. The framework is anchored by a central **Orchestrator Agent**, which manages the overall generation process. The orchestrator interacts with a collection of **Helper Agents**, each specializing in a particular domain or information block as defined in the dependency graph. The orchestrator executes tasks through a step-by-step reasoning process, delegating subtasks to the relevant helper agents as needed and integrating their responses to generate coherent, context-aware outputs. For our experiments, we utilized the GPT-4o model (OpenAI, 2024); however, our approach is model-agnostic and not dependent on any specific architecture.

**The Orchestrator Agent**

The orchestrator is an LLM agent designed using the **ReAct (Reasoning + Action)** paradigm (Yao, 2023). In this architecture, the agent alternates between reasoning steps—where it interprets its current state—and action steps—where it chooses to either query a helper agent, call a predefined function, or revise a previous output based on feedback. Each result is treated as an observation that informs the next reasoning cycle. This iterative strategy leads to greater robustness, enabling the orchestrator to revise its own intermediate conclusions if errors or inconsistencies arise. To guide its behavior, the orchestrator is equipped with a generalizable task strategy that outlines when and how to query each relevant helper agent based on the dependencies of the current information block. This strategy allows the orchestrator to adapt flexibly to different training scenarios and target blocks.

**Helper Agents**

Each Helper Agent is implemented as an LLM equipped with a **Retrieval-Augmented Generation (RAG)** mechanism (Lewis, 2020). These agents are responsible for specific input blocks and are provided with relevant scenario data, typically in the form of structured or semi-structured documents such as JSON files. When queried by the orchestrator, each helper responds precisely to the question asked based solely on its domain-specific information. This modular design allows each helper to function as an expert on a narrow slice of the scenario.

**The Map and MCOO Helper Agent**

Among the specialized helper agents, the Map and MCOO (Modified Combined Obstacle Overlay) Helper Agent is unique in that it supports visual context generation for geospatial reasoning tasks. This agent serves as a tool-augmented utility capable of producing tailored visualizations of terrain, unit positions, movement corridors, and obstacles in response to orchestrator queries.

Earlier attempts to describe this information purely in text led to poor results—LLMs often misinterpreted spatial layouts, confused directions, or failed to relate terrain features to movement planning. To address this issue, the Map and MCOO Helper Agent uses programmatic tools to generate simplified visual overlays of the operational environment, which are then provided as inputs to the Orchestrator Agent. These visual representations enable the orchestrator to make spatial inferences using its multimodal capabilities.

Importantly, the helper does not pass full map images to the orchestrator. Large, high-resolution images (Fig. 2a) tend to overwhelm multimodal LLMs with irrelevant noise. Instead, the helper extracts and renders only the relevant elements—such as specific locations, terrain features, obstacles, and phase lines—based on the orchestrator's query context. This filtered, focused rendering improves the orchestrator's ability to reason accurately about space and movement (Fig. 2b).

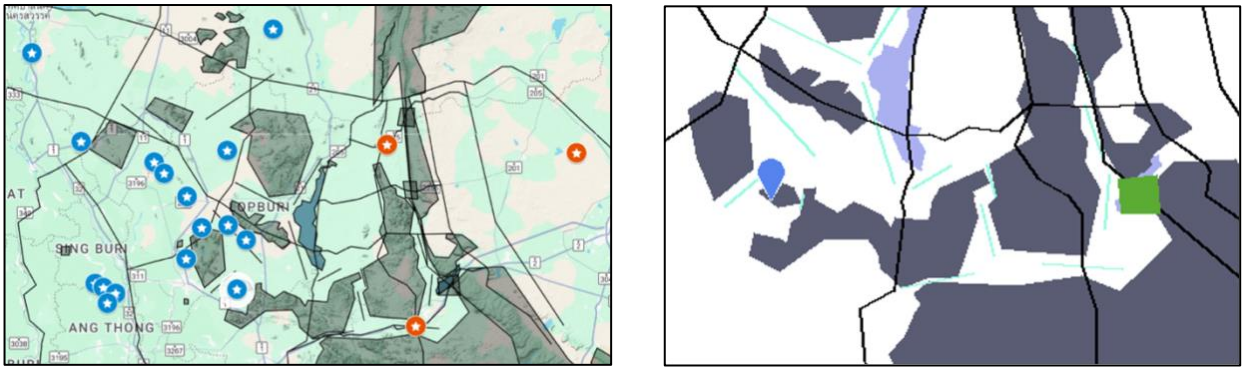

**Fig 2. (a) (Left) MCOO representation of Terrain map of Area of Operation including Phase Lines, Unit Placements, Obstacles, and Movement Corridors. (b) (Right) Simplified representation of the MCOO with a focus on specific objects, here 3DIV and OBJ JAGUARS, and highlighted obstacles and movement corridors**

We have further enhanced this agent with several capabilities to support dynamic movement reasoning. These include:

- A **discretized, waypoint-based movement system** that abstracts the map into traversable nodes, effectively generating a graph and enabling graph-based reasoning for path generation.
- A **multi-path generation function** that identifies multiple viable routes between any source and destination, offering the orchestrator options to evaluate and compare.
- A **progress-tracking mechanism** that visualizes intermediate path points along a selected route, allowing the orchestrator to assess how much of the route has been completed—useful for generating time-based reasoning or movement updates.

Through this functionality, the Map and MCOO helper agent serves not only as a source of map information, but also as an **interactive visual tool provider**, tightly integrated into the orchestrator's reasoning loop.

**PROOF OF CONCEPT GENERATION EXAMPLES**

With the framework in place, we now turn to the technical details of how this architecture generates the two selected example blocks: **Time-Based Unit Positions** and the **Scheme of Movement and Maneuver** section of the OPORD.

**Predicting Unit Positions Over Time**

The Unit Positions Time Based block involves predicting the geographic coordinates of military units at specific future time points, given their previously known positions. By forecasting the movements of each participating unit, we can reconstruct the full spatiotemporal layout of forces over the course of the scenario. This block is categorized as Orange, indicating that it requires substantial reasoning and is likely to benefit from human verification and correction. Fig. 3 shows the relevant information blocks from the scenario dependency hierarchy, each of which is assigned as a helper agent in the generation process.

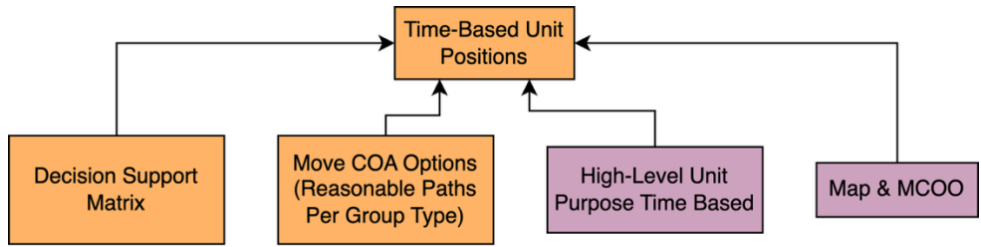

**Fig. 3. Simplified Hierarchy of Unit Positions Time Based**

To generate a unit's predicted position at a future time step, the **multi-agent LLM framework** operates as follows:

1. **Establish Initial Context**: The Orchestrator Agent queries the **High-Level Unit Purpose** helper agent to understand the unit's assigned tasks and objectives. It also retrieves the most recent known location of the unit. In parallel, it queries the **Decision Support Matrix** helper agent to determine if any pre-defined trigger events apply to the unit, which could modify or accelerate its planned course of action.
2. **Determine Goal Location**: Using the objectives and any relevant decision triggers, the orchestrator infers a goal location that the unit is expected to reach after a defined period (e.g., several days).
3. **Select a Movement Path**: To predict intermediate positions, the orchestrator must reason about the route taken between the unit's current and goal locations. It queries the **MCOO helper agent**, which returns a **simplified visual representation** of the terrain, obstacles, movement corridors, and phase lines relevant to the unit's path. The orchestrator agent, equipped with multimodal capabilities, analyzes this image and reasons about the optimal path. Importantly, the agent does not always select the shortest route. It incorporates contextual reasoning—e.g., avoiding terrain that may be vulnerable to ambush or difficult to traverse—to make tactically sound movement predictions.
4. **Estimate Progress and Interpolate Positions**: Once a movement path has been selected, the orchestrator calls the **MCOO helper agent** again to generate a **waypoint-based visualization** of the route. This enables the orchestrator to estimate progress along the path at a given time step and thereby interpolate the unit's expected coordinates. For example, if a unit is expected to travel a path over five days, the orchestrator can estimate the unit's position on Day 3 based on partial path completion, terrain difficulty, and any relevant operational constraints.

Figure 4 presents a simulated example of our multi-agent framework executing a reasoning task. In this trace, the Orchestrator Agent progresses through a sequence of reasoning. At each step, the agent produces a "Thought" representing its intermediate analysis based on prior observations and internal reasoning. Following this, the agent specifies an "Action," which indicates the helper agent to be invoked and the functional capability to be accessed. The accompanying "Action Input" is a structured parameter set generated by the Orchestrator to direct the helper agent's response.

Upon executing the action, the helper agent returns an "Observation"—a response that provides the required information or result. The Orchestrator then integrates this observation into its context and proceeds to the next reasoning step. The framework supports recursive control: if the Orchestrator determines that an action has produced an unsatisfactory or inconsistent outcome, it is capable of backtracking and reissuing a corrected or alternative query. This iterative process continues until the agent determines that a sufficiently complete and consistent final output has been reached.

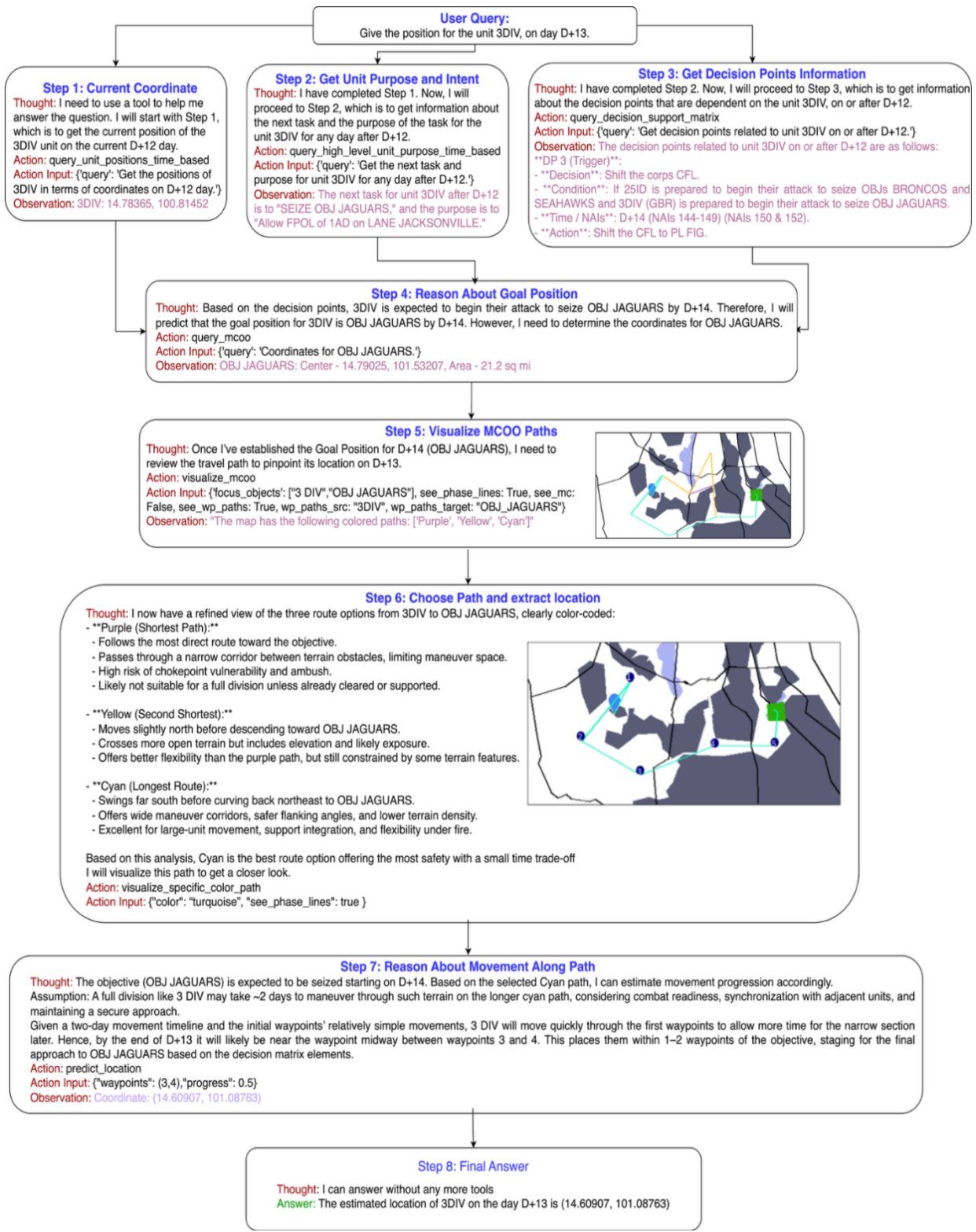

**Fig. 4.** Flowchart describing an example run of the LLM framework. Thoughts, Actions, and Action Inputs are generated by the LLM, and Observations are provided by a helper based on the action

**Generating OPORD Section: Scheme of Movement and Maneuver**

To generate the Scheme of Movement and Maneuver section of the Operations Order (OPORD), we again employ the multi-agent LLM framework described earlier. Unlike previous blocks, this task involves producing final doctrinal text that adheres to military formatting standards. This block is categorized as Green, meaning that it primarily involves synthesizing and rephrasing existing information with limited ambiguity or deep reasoning. Fig. 5 shows the information blocks from the scenario dependency hierarchy relevant to generating this section of the OPORD.

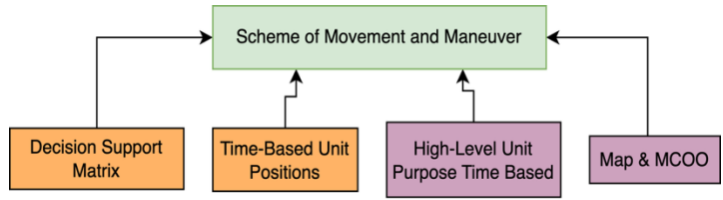

Fig 5. Simplified Hierarchy of Scheme of Movement and Maneuver

The orchestrator follows a structured, multi-step approach that closely resembles the logic used in generating Unit Positions Time Based:

1. **Extract Unit Intent and Tasks:** The orchestrator begins by querying the High-Level Unit Purpose agent to understand the intended role and mission of each unit.
2. **Incorporate Time-Based Positions:** It then retrieves projected unit movements from the Unit Positions Time Based helper, providing spatial-temporal context for each unit's actions.
3. **Align Movements with Triggers:** The orchestrator cross-references the Decision Support Matrix to identify any conditional behaviors or contingency actions associated with each unit. This ensures the narrative accounts for decision points that may affect movement synchronization or task execution.
4. **Align Movements with Map and MCOO information**: The orchestrator can gather information regarding the movement directions for each unit from the MCOO helper. It can also help in gathering knowledge about the phase lines and their ordering for giving more descriptive paragraphs.

Using the combined information, the orchestrator composes a cohesive paragraph for the OPORD section that describes the intended movement and maneuver of all relevant units. The emphasis is on describing coordinated action—how units relate to each other in space and time—rather than generating isolated descriptions. Below, we have shown an example of the OPORD text section as generated by the LLM framework and how it compares to the OPORD created by a human expert.

Fig 6. Scheme of Movement and Maneuver LLM Generated vs Human Expert

**DISCUSSION**

The current process of developing scenarios and generating associated key documents and artifacts for military training remains laborious, resource-intensive, and largely manual. Despite the availability of various tools, existing automated models are not yet capable of producing large-scale scenarios that are both operationally complex and adaptable to dynamic needs. In this context, generative language models—particularly transformer-based large language models (LLMs)—offer a promising avenue for supporting human decision-makers in streamlining and enhancing scenario development.

Recent advancements in LLM applications across the military domain underscore this potential. For instance, Goecks and Waytowich (2024) demonstrated that their COA-GPT system significantly accelerates the development of Courses of Action (COAs) while improving alignment with commander intent. Similarly, Smartbooks (Reddy et al., 2023) presents a system that analyzes open-source news streams and generates temporally structured summaries as question-answer pairs, facilitating rapid understanding of complex information environments. An emerging vision for AI-augmented Command and Control (C2) systems emphasizes dynamic human-AI teaming. In such systems, AI is not only responsible for initial generation of COAs but also supports iterative refinement in concert with human planners (Madison et al., 2024). In follow-up work, Madison et al. (2025) highlighted the importance of maintaining meaningful human control to ensure situational awareness and human authority over final decisions. Complementing these approaches, Van Roo (2024) advocates for incorporating structured reasoning mechanisms to serve as orchestrators—tools that align LLM outputs with mission objectives while providing transparent, traceable reasoning pathways.

Building on these insights, our prior work has shown that LLM-based systems, when supported by an appropriate reasoning structure, can successfully coordinate and synthesize information from diverse sources (Li, Hans, & Ustun, 2025). In this paper, we extend this line of inquiry by applying LLMs to the automation of scenario generation for large-scale combat operations. In collaboration with subject matter experts at Army University, we developed a pipeline that leverages LLMs to generate training artifacts, thereby reducing manual effort and compressing the scenario development timeline.

Our proposed multi-agent LLM-based framework can generate scenario artifacts from both unstructured textual inputs and structured geospatial data, such as unit locations and terrain constraints. Recent integration of a 2D simulation environment further enhances the dynamic movement reasoning capabilities of this pipeline by enabling automatic analysis of terrain features—such as identifying avenues of approach—directly from geospatial layouts. This framework has shown particularly strong performance in tasks like generating Schemes of Movement and Maneuver section of an Operation Order. In these tasks, the system effectively maps unit motivations to geospatial data and engages in detailed reasoning to generate operationally relevant outputs. Central to this capability is our multi-agent, multi-modal approach, which decomposes reasoning into a step-by-step process that improves coherence, supports intermediate validation, and increases the reliability of final outputs.

**CONCLUSION**

Mastering complex military planning tasks requires more than computational power—it requires structured reasoning, adaptability, and the ability to synthesize diverse operational perspectives into coherent strategies. While traditional AI systems excel at optimizing for specific, isolated objectives, they often fail to reconcile conflicting constraints or adapt flexibly to novel conditions. Large Language Models (LLMs), particularly generative transformers, offer promising capabilities for supporting scenario development processes. However, in standard single-agent prompting settings, these models rely on linear text generation, which can hinder their ability to maintain contextual consistency and handle iterative reasoning.

In contrast, this paper discusses the utility of multi-agent, multi-modal LLM frameworks as a more effective paradigm. By distributing tasks across specialized agents, this approach mirrors the way human teams tackle complex problems. It allows for modular reasoning, structured workflows, and coordinated refinement of scenario artifacts. Our results demonstrate that LLMs, when guided by structured reasoning and supported by geospatial context, can play a meaningful role in automating and augmenting military scenario development, underscored by our success in generating coherent and structured scenario elements for large-scale operations.

**ACKNOWLEDGEMENTS**

The authors acknowledge the use of Large Language Models for assistance with proofreading and grammar checking. All content was reviewed, edited, and approved by the human authors, who take full responsibility for the final manuscript. The project or effort depicted was or is sponsored by the U.S. Army Combat Capabilities Development Command -- Soldier Centers under contract number W912CG-24-D-0001. The content of the information does not necessarily reflect the position or the policy of the Government, and no official endorsement should be inferred.